\documentclass[conference,a4paper]{IEEEtran}

\usepackage[utf8]{inputenc} 
\usepackage[T1]{fontenc}
\usepackage{url}              
\usepackage{cite}             
\usepackage{style}
\usepackage{amssymb}
\usepackage{mleftright}       
\mleftright                   
\usepackage{wrapfig}
\usepackage{graphicx}         
\usepackage{booktabs}         

\usepackage[ruled,vlined,linesnumbered]{algorithm2e}
                              
\begin{document}

\title{Network Fault-tolerant and Byzantine-resilient Social Learning via Collaborative Hierarchical Non-Bayesian Learning  
} 

\author{Connor Mclaughlin*, Matthew Ding*, Deniz Edogmus, and Lili Su
\thanks{*student authors with equal contribution.\\ 
This research was supported by ONR award N00014-18-9-0001. }
\thanks{C.\,Mclaughlin, D.\,Edogmus, and L.\,Su are with Electrical and Computer Engineering Department, Northeastern University, Boston (emails: \{mclaughlin.co, d.erdogmus, l.su\}@northeastern.edu)}
\thanks{M.\,Ding is with Department of Electrical Engineering and Computer Sciences, University of California, Berkeley  (email: matthewding@berkeley.edu)}
}

\maketitle

\begin{abstract}
As the network scale increases, existing fully distributed solutions start to lag behind the real-world challenges such as (1) slow information propagation, (2) network communication failures, and (3) external adversarial attacks. In this paper, we focus on hierarchical system architecture and address the problem of non-Bayesian learning over networks that are vulnerable to communication failures and adversarial attacks. 

On network communication, we consider packet-dropping link failures. 
We first propose a hierarchical robust push-sum algorithm that can achieve average consensus despite frequent packet-dropping link failures. We provide a sparse information fusion rule between the parameter server and arbitrarily selected network representatives. Then, interleaving the consensus update step with a dual averaging update with Kullback–Leibler (KL) divergence as the proximal function, we obtain a packet-dropping fault-tolerant non-Bayesian learning algorithm with provable convergence guarantees.  

On external adversarial attacks, we consider Byzantine attacks in which the compromised agents can send maliciously calibrated messages to others (including both the agents and the parameter server). To avoid the curse of dimensionality of Byzantine consensus, we solve the non-Bayesian learning problem via running multiple dynamics, each of which only involves Byzantine consensus with scalar inputs. To facilitate resilient information propagation across sub-networks, we use a novel Byzantine-resilient gossiping-type rule at the parameter server. 

\end{abstract}

\section{Introduction}
\label{sec: intro}
\begin{figure}[h]
\centering
\includegraphics[width=0.4\textwidth]{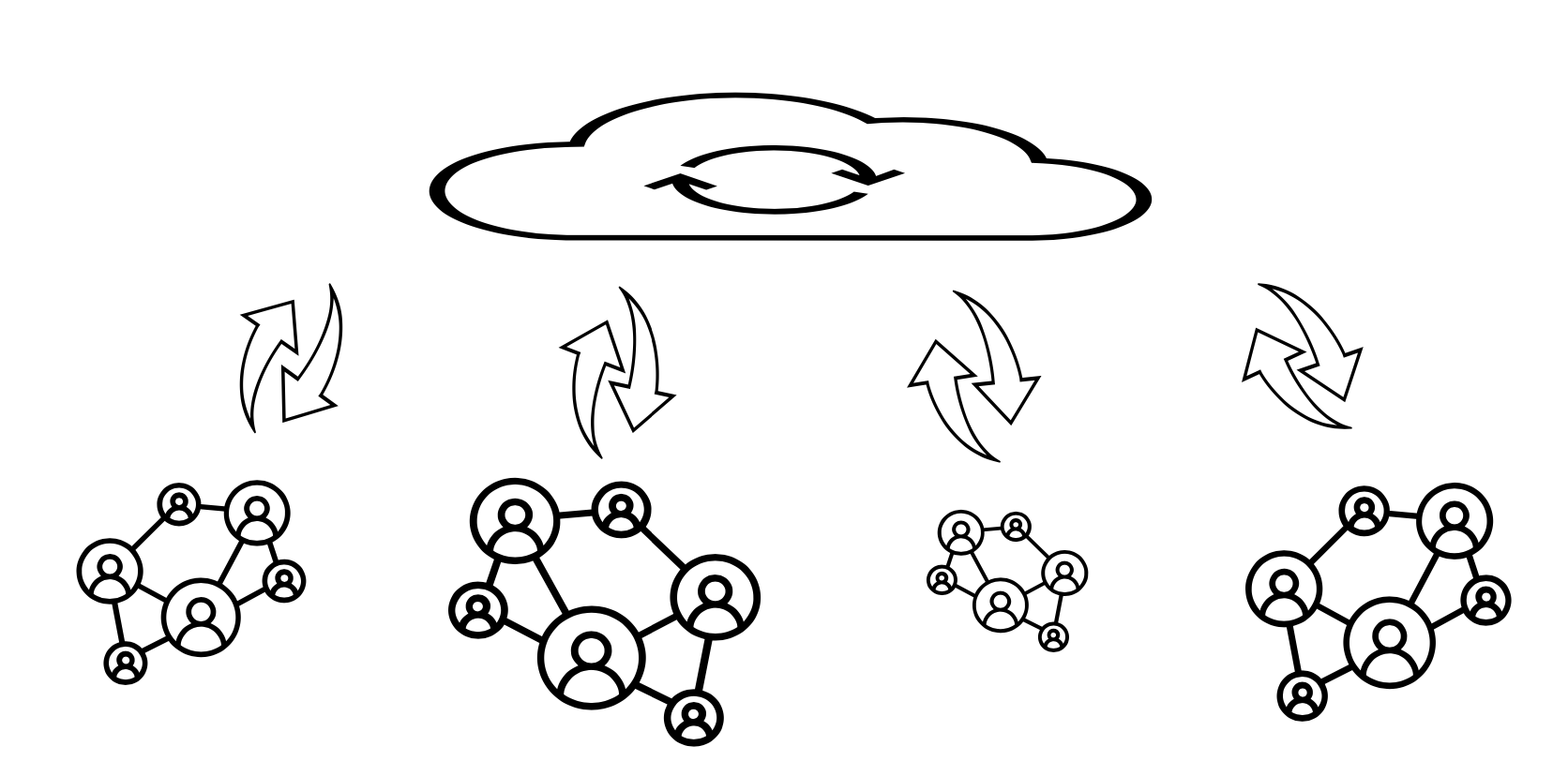}
\caption{A hierarchical system architecture}
\label{fig: hierarchical architecture}
\end{figure}

As the scale of the multi-agent network increases, existing fully distributed solutions start to lag behind the crucial real-world challenges such as (1) slow information propagation, (2) network communication failures, and (3) external adversarial attacks. Towards scalable decentralized solutions, instead of a gigantic multi-agent network, we consider a hierarchical system architecture in which the agents are clusters into $M$ sub-networks, and a parameter server exists to aid the information exchanges among sub-networks. The system architecture is depicted in Fig.\ref{fig: hierarchical architecture}. 
Similar system architecture is adopted in the literature~\cite{edge1, edge2,Hierarchical_FL} 
Sending messages between an agent and the parameter server is costly; hence needs to be sparse.  
In this paper, we addresses the problem of hierarchical non-Bayesian learning over the multi-agent networks that are vulnerable to communication failures and Byzantine attacks. 
 Non-Bayesian learning \cite{jadbabaie2012non,jadbabaie2013information,molavi2017foundations,nedic2015nonasymptotic} is a ``consensus + innovation'' approach. It is a computational efficient approximation to Bayesian learning over networks wherein the information is scattered over different agents, and there does not exist an agent that can learn the truth by itself. 
Mathematically, social learning can be formulated as a distributed  multiple hypothesis testing problem. Let $\Theta = \{\theta_1, \cdots, \theta_m\}$ be the set of $m$ hypotheses.  
There is an unknown underlying truth $\theta^*\in \Theta$ that determines the joint distribution of the local measurements at individual agents. For any given hypothesis, the marginal distributions at the agents can be different. Moreover, for any given agent, its marginal distributions may be identical under different hypotheses, i.e., ``local confusion'' often exists. The goal of non-Bayesian learning is to design proper consensus and innovation components to enable the agents to collaboratively identify the underlying truth $\theta^*$.

On network failures, we consider the challenging packet-dropping link failures, i.e., a communication link may drop the transmitted messages unexpectedly and without notifying the sender. 
As observed in \cite{RobustAverageConsensus,spiridonoff2020robust,UnreliableConvergence}, this link failure is much harder to tackle compared with the ones wherein each agent is aware of the message delivery status.  
Though assuming knowledge of out-going degree is reasonable \cite{duchi2011dual}, in harsh and versatile deployment environments such as undersea, 
the communication between neighboring entities may suffer strong interference, leading to rapidly changing channel conditions and, consequently, possible unsuccessful message delivery.

On adversarial attacks, we consider Byzantine attacks in which the compromised agents can send maliciously calibrated messages to others (including both the agents and the parameter server). Tolerating Byzantine attacks is challenging \cite{lamport1982byzantine}. Byzantine resilience suffers curse of dimensionality -- no Byzantine consensus algorithms can tolerate more than $\min\{1/3, \,\, 1/(d+1)\}$, where $d$ is the input dimension, fraction of Byzantine agents for synchronous systems. 
%

\vskip \baselineskip
\noindent{\bf Contributions.}
Our contributions are two-fold: 
\begin{itemize}
    \item We first consider link failures. We propose a hierarchical push-sum (HPS) algorithm that can achieve average consensus despite frequent packet-dropping link failures. A key algorithmic novelty is the design of a sparse information fusion rule between the parameter server and arbitrarily selected network representatives. 
    Interleaving the HPS update with a dual averaging update with Kullback–Leibler (KL) divergence as the proximal function (i.e.\, the innovation step), we obtain a packet-dropping fault-tolerant non-Bayesian learning algorithm with provable convergence guarantees.  
    \item Then we consider Byzantine attacks. We propose an efficient algorithm that is resilient to arbitrary placement of $F$ Byzantine agents provided that $M\ge F+1$ and there exists at least $F+1$ subnetworks each of which contains $<1/3$ fraction Byzantine agents.\footnote{Formal description of the conditions can be found in Section \ref{sec: Byzantine non-Bayesian}.}   
    We solve the non-Bayesian learning problem via running multiple dynamics, each of which only involves Byzantine consensus with scalar inputs. To facilitate resilient information propagation across sub-networks, we use a novel Byzantine-resilient gossiping-type rule at the parameter server. Under mild technical assumptions, we show that this algorithm enables each normal (i.e.\,non-Byzantine) agent to identify $\theta^*$ with probability 1 for any finite fusion frequency with the parameter server. 
\end{itemize}

\section{System and Threat Models}
\subsection{System Model}
The system consists of a parameter servers and $M$ sub-networks. 
The connection among each multi-agent network $S_i$ is time-varying and is formally represented by  graphs $G(\calV_i, \calE_i[t])$, where $\calV_i=\{v^i_1, \cdots, v^i_{n_i}\}$ is node set and $\calE_i[t]$ is the set of all directed edges. There exists $\calE_i$ such that $\calE_i[t]\subseteq \calE_i$ for each $t$.   Let $N:=\sum_{i=1}^M n_i$.  
Agents in the same sub-network can exchange messages subject to the given communication network $G(\calV_i, \calE_i[t])$ at time $t$. No messages can be exchanged directly between agents in different sub-networks. In addition,  the PS has the freedom in querying and pushing messages to any agent. Nevertheless, such message exchange is costly and needs to be sparse. 

For an arbitrary agent $j$ in network $S_i$, 
let $\calI_j^i[t] = \{k \mid (k,j) \in \calE_i[t] \}$ 
and $\calO_j^i[t] = \{k \mid (j,k) \in \calE_i[t] \}$, respectively, be the sets incoming and outgoing neighbors to agent $j$. For notational convenience\footnote{This will not create confusion in our analysis because only $\abth{O_j^i[t]}$ is used in the algorithm.}, we denote $d_j^i[t] = \abth{O_j^i[t]}$.

\subsection{Threat Model}
\label{subsec: threat model} 
\paragraph{Packet-dropping failures}
We follow the network fault model adopted in \cite{UnreliableConvergence}. 
Specifically, any communication link may unexpectedly drop a packet transmitted through it, and the sender is unaware of such packet lost. 
If a link successfully deliver messages at communication round $t$, we say this link is {\em operational} at round $t$.  
We assume that for each $i\in [M]$, any link $(j,j^{\prime})\in \calE_i$ is operational at least once every $B$ iterations. 

Essentially, since we allow the communication networks to be time-varying, our network faults model is more general than the link failures considered in \cite{duchi2011dual} wherein the outgoing degree is known.

\paragraph{Byzantine faults}
We adopt Byzantine fault model \cite{Lynch:1996:DA:2821576,lamport1982byzantine} -- a canonical fault model in distributed computing.  

There exists a system adversary that can choose up to $F$ of the $N$ agents (where $F< N$) to compromise and control.  
An agent suffering Byzantine fault is referred to as Byzantine agent.  Let $\calA \subset \cup_{i=1}^M\calV_i$ such that $\abth{\calA} \le F$ be the {\it unknown} subset of $\cup_{i=1}^M\calV_i$ that contains all the Byzantine agents.  
We assume that each non-Byzantine agent knows $F$, which is a standard assumption in the literature \cite{Lynch:1996:DA:2821576}. The system adversary is very {\it powerful} in the sense that it has complete knowledge of the system, including the local program that each good agent is supposed to run and the problem inputs. 
The Byzantine agents can collude with each other and deviate from their pre-specified local programs to {\em arbitrarily} misrepresent information to the good agents with the only restriction that the communication channel is authenticated, i.e., a Byzantine  agent cannot forge the digital signature of someone else.  
Moreover, Byzantine agents can use point-to-point rather than broadcast communication. Formally, let $m_{jj_1}(t)$ and $m_{jj_2}(t)$ be the messages sent by agent $j$ to two distinct outgoing neighbors $j_1$ and $j_2$. 
Under point-to-point communication, it is allowed that 
\[
m_{jj_1}(t) \not= m_{jj_2}(t). 
\]

\begin{remark}[Lower bound of Byzantine resilience]
No consensus algorithms can tolerate $\ge 1/3$ of agents to be Byzantine even in the simple setting where the graph is complete and inputs are binary $\{0, 1\}$. 
In general, when the inputs are multi-dimensional, i.e., $d\ge 2$, the tolerable fraction of Byzantine agents can be much less than $1/3$. 
The following lower bound (impossibility) results are derived in \cite{mendes2015multidimensional}: 
There {\bf does not} exist Byzantine consensus algorithm that can tolerate Byzantine fraction to be 
\begin{align*}
F/N > 
\begin{cases}
\min\{1/3, \,\, 1/(d+1)\}, &\text{for synchronous systems; }\\   
1/(d+2), & \text{for asynchronous systems.}
\end{cases}
\end{align*}
\end{remark}
Fortunately, we are able to solve a $d$-dimensional non-Bayesian learning problem via a collection of scalar problems.

\section{Social Learning Problem}
\label{sec: problem formulation}
%
%
%
We following a canonical learning model in social networks/multi-agent systems 
\cite{jadbabaie2012non,jadbabaie2013information,nedic2015nonasymptotic}. 
The entire system can be in one of the $m$ possible unknown environments $\Theta = \{\theta_1, \theta_2, \cdots, \theta_m\}$. Let $\theta^* \in \Theta$ denote the underlying environment that the normal agents try to collaboratively learn based on their locally collected signals. 

For each time $t$, each agent {\em independently} obtains private signal about the environmental state $\theta^*$, which is initially unknown to every agent in the network.
For ease of exposition, we assume that if multiple signals are observed, only one signal is used to update beliefs.
We use $i_j$ to denote the $j$-th agent in the $i$-th network $S_i$.  
Each agent $i_j$ knows the structure of its private signal, which is represented by a collection of parameterized distributions $\calD^{i_j}=\{\ell_{i_j}(w_{i_j} | \theta)| \theta\in \Theta,\, w_{i_j}\in \calS_{i_j}\}$,
where $\ell_{i_j}(\cdot | \theta)$ is a distribution with parameter $\theta\in \Theta$, and $\sup_{w_{i_j}\in \calS_{i_j}, \text{and }\theta, \theta^{\prime}\in \Theta} \log\frac{\ell_{i_j}(w_{i_j} | \theta)}{\ell_{i_j}(w_{i_j} | \theta^{\prime}}\le L$ for some positive constant $L>0$. 
Precisely, let $s_t^{i_j}$ be the private signal observed by agent ${i_j}$ in iteration $t$, and let ${\bf s}_t=\{s_t^{1}, \cdots, s_t^{N}\}$ be the signal profile at time $t$ (i.e., signals observed by the agents in iteration $t$). Given an environmental state $\theta$, the signal profile ${\bf s}_t$ is generated according to the joint distribution $\bm{\ell}(\cdot\mid\theta) = \ell_{1}(\cdot|\theta)\times \cdots \times \ell_{N}(\cdot|\theta)$.

\section{Average Consensus in the Presence of Packet-dropping Failures}
\label{sec: average consensus}
In \cite{UnreliableConvergence}, we proposed a fast robust push-sum algorithm that can achieve average consensus on a single network. In this section, we extend our prior work to the hierarchical system architecture, formally described in Algorithm \ref{alg:push-sum hierarchical FL}. 
Up to line 11 is the parallel execution of the fast robust push-sum \cite{UnreliableConvergence} over the $M$ subnetworks. Lines 12-21 describes the novel information fusion cross the $M$ subnetworks, which only occurs every other $\Gamma$ iterations. 
\begin{algorithm}
\caption{Hierarchical Push-sum (HPS)}
\label{alg:push-sum hierarchical FL}
{\small {\em Initialization}: 
For each sub-network $i=1, \cdots, M$: $z_j^i[0]=w_j^i\in R^d$, $m^i_j[0]=1\in R,$ 
$\sigma_j^i[0]={\bf 0}\in R^d$, $\tilde{\sigma}_j^i[0]=0\in R$, and $\rho_{j^{\prime}j}[0]={\bf 0}\in R^d$, $\tilde{\rho}_{j^{\prime}j}[0]=0\in R$ for each incoming link, i.e., $j^{\prime} \in \calI_j^i$.

\vskip 0.2\baselineskip 
In parallel, each client $j\in \calV_i$ in parallel does:\\
\For{$t\ge 1$}
{ 
$\sigma_j^{i+}[t]  \gets  \sigma_j^i[t-1] + \frac{z_j^i[t-1]}{d_j^i[t]+1}$,
$\tilde{\sigma}_j^{i+}[t] \gets \tilde{\sigma}_j^i[t-1] + \frac{m_j^i[t-1]}{d_j^i[t]+1}$\;

Broadcast $\pth{\sigma^{i+}_j[t], \tilde{\sigma}^{i+}_j[t]}$ to outgoing neighbors\;

\For {each incoming link $(j^{\prime},j)\in \calE_i$}
{\eIf{message $\pth{\sigma^{i+}_{j^{\prime}}[t], \tilde{\sigma}^{i+}_{j^{\prime}}[t]}$ is received}
{$\rho^i_{j^{\prime}j}[t] \gets \sigma^{i+}_{j^{\prime}}[t]$, ~~ $\tilde{\rho}^i_{j^{\prime}j}[t] \gets \tilde{\sigma}^{i+}_{j^{\prime}}[t]$\;}
{ $\rho^i_{j^{\prime}j}[t] \gets \rho^i_{j^{\prime}j}[t-1]$, ~~$\tilde{\rho}^i_{j^{\prime}j}[t] \gets \tilde{\rho}^i_{j^{\prime}j}[t-1]$\;}
$ z_j^{i+}[t] \gets \frac{z_j^{i}[t-1]}{d_j^i[t]+1} +  \sum_{j^{\prime}\in \calI_j^i} \pth{\rho_{j^{\prime}j}[t] - \rho_{j^{\prime}j}[t-1]}$, 
$m_j^{i+}[t]  \gets \frac{m_j^i[t-1]}{d_j^i[t]+1} + \sum_{j^{\prime}\in \calI_j^i} \pth{\tilde{\rho}_{j^{\prime}j}[t] -\tilde{\rho}_{j^{\prime}j}[t-1]}$.
}

$\sigma^i_j[t]  \gets  \sigma^{i+}_j[t] + \frac{z_j^{i+}[t]}{d_j^i[t]+1}$,
$\tilde{\sigma}^i_j[t]  \gets  \tilde{\sigma}^{i+}_j[t] + \frac{m_j^{i+}[t]}{d_j^i[t]+1}$,
$z_j^i[t]  \gets \frac{z_j^{i+}[t]}{d_j^i[t]+1}$,
$m_j^i[t] \gets \frac{m_j^{i+}[t]}{d_j^i[t]+1}$\; 
}

\If{$j$ is a designated agent of network $S_i$}
{
\If{$t\mod \Gamma =0$}
{
Send $\frac{1}{2}z_j^i[t]$ and $\frac{1}{2}m_j^i[t]$ to the PS\; 

Upon receiving messages from the PS {\bf do} \\
update  
$z_j^{i}[t]\gets \frac{1}{2}z_j^{i}[t] + \frac{1}{2M}\sum_{i=1}^M z_{i_0}^i[t]$\; 
$m_j^{i}[t]\gets \frac{1}{2}m_j^{i}[t] + \frac{1}{2M}\sum_{i=1}^M m_{i_0}^i[t]$\;
}
} 

\If{$t\mod \Gamma =0$}
{
The PS does the following: 

Wait to receive $z_{i_0}^i[t]$ and $m_{i_0}^i[t]$ from each designated agent of the $M$ networks\; 

Compute and send $\frac{1}{M}\sum_{i=1}^M \frac{1}{2}z_{i_0}^i[t]$ and $\frac{1}{M}\sum_{i=1}^M \frac{1}{2}m_{i_0}^i[t]$ to all designated agents $i_0$ for $i=1, \cdots, M$. 
} 

}
\end{algorithm}
Similar to the standard Push-Sum \cite{kempe2003gossip}, in addition to the primary variable  $z_j^i$, each agent $j$ keeps a mass variable $m_j^i$ to correct the possible bias caused by the graph structure, and uses the ratio $z_j^i/m_j^i$ to estimate the average consensus. The correctness of push-sum relies crucially on its mass preservation, i.e., $\sum_{i=1}^M\sum_{j=1}^{n_i} m_j^i[t] = N$ for all $t$. The variables $\sigma$, $\tilde{\sigma}$, $\rho$, and $\tilde{\rho}$ are introduced to recover the dropped messages and mass. 
Specifically, $\sigma_j^i[t]$ and $\tilde{\sigma}_j^i[t]$ are used to record how much value and mass have been sent to each of the outgoing neighbor of agent $j$ up to time $t$. Corresponding, $\rho_{j^{\prime}j}[t]$ and $\tilde{\rho}_{j^{\prime}j}[t]$ are used to record how much value and mass have been received by agent $j$ through the link $(j^{\prime}j)$. To control the trajectory smoothness of the $z_j^i/m_j^i$, in each iteration, both $z$ and $m$ are updated twice. 

For each network, we choose an arbitrary agent as the network representative, and only this designated agent can exchange messages with the PS. Let $i_0$ denote the designated agent of network $i$. 
Every other $\Gamma$ iterations, each designated agent pushes 1/2 of its local value and mass to the PS. The PS computes the received average estimate and mass, and sends the averages back to each designated agent. Each designated agent then updates its local estimates and mass as ones pushed back from the PS.  

\begin{assumption}
\label{ass: connectivity}
Each network $(\calV_i, \calE_i)$ is strongly connected for $i=1, \cdots, M$. 
\end{assumption}

Denote the diameter of $G(\calV_i, \calE_i)$ as $D_i$. Let $D^* := \max_{i\in [M]}D_i$.  
Let $\beta_i= \frac{1}{\max_{j\in \calV_i} (d_j^{i}+1)^2}$. 
\begin{theorem}
\label{rps convergence rate}
Choose $\Gamma = BD^*$.  
Suppose that Assumption \ref{ass: connectivity} hold, and that $t\ge 2\Gamma$.
Then 
\begin{align*}
\norm{\frac{z_j^i[t]}{m_j^i[t]} -\frac{1}{N}\sum_{j=1}^{N} w_j^i} 
&\le \frac{4M^2\sum_{j=1}^{N}\norm{w_{j^{\prime}}^i}}{\pth{\min_{i\in [M]}\beta_i}^{2D^*B}N}  \,\gamma^{\lfloor t/2\Gamma \rfloor - 1}, 
\end{align*} 
where $\gamma = 1-\frac{1}{4M^2} \pth{\min_{i\in [M]}\beta_i}^{2D^*B}$. 
\end{theorem}
Henceforth, for ease of exposition, we adopt the simplification that $\lfloor t/2\Gamma \rfloor - \lceil r/2\Gamma \rceil =  (t-r)/2\Gamma.$ 
Such simplification does not affect the order of convergence rate. 
Exact expression can be recovered while a straightforward bookkeeping of the floor and ceiling in the calculation. 

Theorem \ref{rps convergence rate} says that, despite packet-dropping link failures and sparse communication between the networks and the PS, the consensus error $\norm{\frac{z_j^i[t]}{m_j^i[t]} -\frac{1}{N}\sum_{j=1}^{N} w_j^i}$ decays to 0 exponentially fast.  
Clearly, the more reliable the network (i.e. smaller $B$) and the more frequent across networks information fusion (i.e. smaller $\Gamma$), the faster the convergence rate. 
\begin{remark}
Partitioning the agents into $M$ subnetworks immediately leads to smaller network diameters $D^*$. 
Hence, compared with a gigantic single network, the term $\pth{\min_{i\in [M]}\beta_i}^{2D^*B}$ for the $M$ sub-networks is significantly larger, i.e., faster convergence.  
\end{remark}

\begin{remark}
It turns out that our bound in Theorem \ref{rps convergence rate} is loose in quantifying the total number of global communication. Specifically, for any given $\epsilon>0$, to reduce the error to $O(\epsilon)$, based on the bound in Theorem \ref{rps convergence rate}, it takes 
$t\ge \Omega\pth{\Gamma \log \epsilon/\log \gamma}.$
 The total global communication cost is around $\Theta( \log \epsilon/\log \gamma)$ -- hinting that less frequent communication does not save global communication. However, our preliminary simulation and experiment results show that, up to certain region, less frequent (i.e., large $\Gamma$) communication does not lead to increase of training error. 
\end{remark}

\vskip 0.5\baselineskip
The analysis of Theorem \ref{rps convergence rate} relies on a construction of augmented graphs and a compact matrix representation of the dynamics of $z$ and $m$ over those augmented graphs.  Since the update of value $z$ and weight $m$ are identical, 
henceforth, we focus on the value sequence $z$. Let $\tilde{N}$ denote the number of vertices of the augmented graph. Let $\bm{z}\in R^{d\tilde{N}}$ be the vector that stacks the local values of each vertex in the {\em augmented graph}.  
We show that $\bm{z}$ evolves as 
\begin{align*}
    \bm{z}[t] = \pth{\bm{M}[t] \otimes \bm{I}}\bm{z}[t-1],
\end{align*} 
where $\otimes$ denotes the Kronecker product, and $\bm{M}[t]$ is a stochastic matrix that captures the mutual influences of the agents.  
The fact that $\bm{M}[t]$ is time-varying is because the link status is time-varying. 
For each $t\mod \Gamma \not=0$, $\bm{M}[t]$ is a block matrix with $M$ blocks.  
Fix $t$ be arbitrary iteration such that $t\mod \Gamma =0$. We construct matrix $\bm{M}$ in two steps. We let $\bar{\bm{M}}$ denote the matrix constructed the same way as for $t\mod \Gamma \not=0$. Let $\bm{F}\in \reals^{\tilde{N}\times \tilde{N}}$ be the matrix that captures the mass push among the designated agents under the coordination of the parameter server. Specifically, 
\begin{align*}
\bm{F}_{j_0,j_0} &= \frac{M+1}{2M} ~~~~ \text{for each designated agent } j_0; \\
\bm{F}_{j_0,j^{\prime}_0} & = \frac{1}{2M} ~~~~ \text{for any} j_0\not= j^{\prime}_0,
\end{align*}
with all the other entries being zeros. Henceforth, we refer to matrix $\bm{F}$ as hierarchical fusion matrix. 
Clearly, $\bm{F}$ is a doubly-stochastic matrix. Hence, we define $\bm{M}$ as 
\begin{align}
\label{eq: fusion iteration}
\bm{M}[t] =\bm{F} \bar{\bm{M}}[t]. 
\end{align}
Let ${\bf \Psi}(r,t)\triangleq \prod_{\tau=r}^t\, {\bf M}^{\top}[\tau]={\bf M}^{\top}[r] {\bf M}^{\top}[r+1]\cdots {\bf M}^{\top}[t]$ denote the matrix product, 
where $r\le t$ with ${\bf \Psi}(t+1,t)\triangleq {\bf I}$ by convention. 
%
Notably, ${\bf M}^{\top}[\tau]$ is row-stochastic. 

The following lemma is useful in the analysis of our non-Bayesian learning algorithms. 
Due to space limitation, its proof is omitted. 
\begin{lemma}
\label{lm: entire matrix lower bound}
Let $D^* := \max_{i\in [M]}D_i$. 
Choose $\Gamma = BD^*$. 
Suppose that $t-r+1\ge 2\Gamma$. Then every entry of the matrix product ${\bf \Psi}(r, t)$ is lower bounded by $\frac{1}{4M^2}\pth{\min_{i\in [M]}\beta_i}^{2D^*B}$. 
\end{lemma}

\section{Non-Bayesian Learning: Packet-dropping Links}
\label{sec: algorithm: dropping link}

Let $\Delta_{\Theta}$ denotes the probability simplex over $\Theta$.  
Each agent $j_i$ keeps a local variable $\mu_j^i\in \Delta_{\Theta}$, which we refer to as a {\em belief approximation vector}. Notably, there is a common abuse of terminology of ``belief vector'' in the literature of non-Bayesian learning \cite{jadbabaie2012non,jadbabaie2013information,nedic2015nonasymptotic}. In contrast to Bayesian learning, the belief vectors in non-Bayesian learning are not the posterior distributions.  
We use $\mu_j^i(\cdot, t)\in \Delta_{\Theta}$ to denote the local estimate at agent $j_i$ at the end of iteration $t$. Let $\mu_j^i(\cdot, 0) =\pth{\frac{1}{m}, \ldots, \frac{1}{m}}^{\top}$ for all $i\in [M]$ and $j\in [n_i]$.\footnote{In this paper, every vector considered is column vector.} 

We would like to design an algorithm that enables 
\begin{align}
\mu_j^i(\theta_{\ell}, t) & \to 1, \text{if } \theta_{\ell} = \theta^* \label{eq: truth learning} \\
\mu_j^i(\theta_{\ell}, t) & \to 0, \text{if } \theta_{\ell} \not= \theta^* \label{eq: non truth elimination}. 
\end{align}

As mentioned in Section \ref{sec: intro}, non-Bayesian learning is a ``consensus''+``innovation'' approach \cite{jadbabaie2012non}. In this section, we use Algorithm \ref{alg:push-sum hierarchical FL} as the consensus component and use the dual averaging with KL divergence as the proximal function. 
Specifically, we add the following lines of pseudo code right after line 12 of the {\bf for-loop} in Algorithm \ref{alg:push-sum hierarchical FL}: 
\vskip 0.6\baselineskip
\fbox{\begin{minipage}{20em}
{\em 
Obtain measurement $s_j^i(t)$;\\ 
\For{$\ell=1, \cdots, m$}
{
$z_j^i(\theta_{\ell}, t) \gets  z_j^i(\theta_{\ell}, t) + \log\pth{\ell(s_j^i(t)\mid \theta_{\ell})}$;
}
$\mu_j^i(\cdot, t) \gets \prod_{\mu\in \Delta_{\Theta}}^{\varphi}\pth{\frac{z_j^i(\cdot, t)}{m_j^i(t)}, 1}$;      
} 
\end{minipage}}
\vskip 0.6\baselineskip
\noindent Here, $\prod_{\mu\in  \Delta_{\Theta}}^{\varphi}\pth{x, \alpha}: = \arg\min_{\mu\in  \Delta_{\Theta}}    \sth{-\iprod{x}{\mu} + \frac{1}{\alpha}\varphi(\mu)}$, where 
$\varphi(\mu) = D_{KL}\pth{\mu\| \mu_0}$. 
The complete pseudo code can be found in Appendix \ref{app: algorithm: dropping link}. 

The update of the local approximate belief vector $\mu_j^i(\cdot, t)$ has the following explicit expression: 
\begin{align*}
\mu_j^i(\theta_{\ell}, t) = \frac{\mu_0(\theta_{
\ell}) \exp\pth{\frac{z_j^i(\cdot, t)}{m_j^i(t)}}}{\sum_{\ell=1}^m \mu_0(\theta_{
\ell}) \exp\pth{\frac{z_j^i(\cdot, t)}{m_j^i(t)}}}, ~~~~ \forall ~ \ell=1, \cdots, m,      
\end{align*}

\begin{assumption}
\label{ass: global identifiability}
The true state $\theta^*$ is globally observable. That is, for any pair of distinct $\theta$ and $\theta^{\prime}$ in $\Theta$, 
\[
\min_{\theta, \theta^{\prime}\in \Theta ~\text{s.t.}~ \theta \not=\theta^{\prime}} D_{KL}\pth{\bm{\ell}(\cdot \mid \theta^{\prime}) \| ~\bm{\ell}(\cdot \mid \theta)} >0, 
\]
where $D_{KL}$ is the Kullback-Liebler (KL) divergence between two probability distributions\footnote{Let $p$ and $q$ be two distributions 
with a common support,  $D_{KL}(p\|q) := \sum_{k=1}^{\infty}p_k\log \frac{p_k}{q_k}$ for finite (or countable) support.}.
\end{assumption}

\begin{theorem}
\label{thm: iterative}
Suppose that Assumptions \ref{ass: connectivity} and \ref{ass: global identifiability} hold. 
Choose $\Gamma = BD^*$. Suppose that $t\ge 2\Gamma$. For any given $\delta\in (0,1)$, with probability at least $1-\delta$:  For all $\theta\in \Theta\setminus\{\theta^*\}$
\begin{align*}
\log \frac{\mu_j^i(\theta, t)}{\mu_j^i(\theta^*, t)}
& \le  -\frac{t}{N} D_{KL}\pth{\theta^*\|  \theta}   + L\sqrt{2t\log \frac{m}{\delta}}\\
& \qquad + \frac{ 8M^2L \gamma^{\frac{1}{2\Gamma}}}{N \pth{1-\gamma^{\frac{1}{2\Gamma}}}\pth{\min_{i\in [M]}\beta_i}^{2D^*B}}. 
\end{align*} 
\end{theorem}
The first term  goes to $-\infty$ linearly in $t$, the second term arises from the randomness in the local signals, and the last term is due to the cumulative consensus error over time. 

%

\section{Non-Bayesian Learning: Byzantine Resilience}
\label{sec: Byzantine non-Bayesian}
%
\begin{definition}\cite{DBLP:journals/corr/abs-1202-6094,vaidya2012matrix}
\label{def: reduced graph}
Given a graph $G(\calV, \calE)$, a reduced graph is constructed as follows: 
(1) remove all faulty nodes $\calA$, 
(2) remove all the links incident on the faulty nodes $\calA$, and 
(3) for each non-faulty node, remove $F$ additional incoming links. If there are less than $F$ such links, remove all the links. 
\end{definition}

Let $\calG_{\text{info}}$ be the collection of all the information flow graph networks of a given graph $G(\calV, \calE)$. Let 
\begin{align}
\label{eq: reduced graphs number}
\chi_i := |\calG_{\text{info}}(G(\calV_i, \calE_i))|.  
\end{align}

When the inputs are scalars, the following condition is shown \cite{DBLP:journals/corr/abs-1202-6094} to be both necessary and sufficient on the network topological structure for Byzantine-resilient consensus to be achievable on the given network $G(\calV, \calE)$. 
\begin{assumption}
\label{ass: tight condition: single network}
Given a communication graph $G(\calV, \calE)$, each of the reduced graph of $G(\calV, \calE)$, defined as per Definition \ref{def: reduced graph}, contains exactly one source component. 
\end{assumption}
\begin{assumption}
\label{ass: Byzantine: source identifiability}
Suppose that graph $G(\calV, \calE)$ satisfies Assumption \ref{ass: tight condition: single network}. 
For any $\theta\not=\theta^*,$ and for any information flow graph $\calH$ of $G(\calV, \calE)$ with $\calS_{\calH}$ denoting the unique source component, the following holds
\begin{align}
\label{failure identify}
\sum_{j\in \calS_{\calH}} D\pth{\ell_j(\cdot |\theta^*)\parallel\ell_j(\cdot |\theta)}~\not=~0.
\end{align}
\end{assumption}
%
Su and Vaidya \cite{su2019defending} mentioned that when $G(\calV, \calE)$ satisfies Assumptions \ref{ass: tight condition: single network} and \ref{ass: Byzantine: source identifiability} 
every normal agent learns $\theta^*$. Though the idea is interesting, formal analysis is missing.

\vskip \baselineskip
Due to the curse of dimensionality of Byzantine resilience, we can not directly plug in a Byzantine consensus algorithm to serve as the ``consensus'' component. 
In Algorithm \ref{alg: hierarchical: pairwise}, we run, in parallel, multiple linear dynamics, 
wherein $\calC\subseteq \{1, 2, \cdots, M\}$ is the set of networks that satisfy Assumptions \ref{ass: tight condition: single network} and \ref{ass: Byzantine: source identifiability}.  
To restrain the negative impacts of the Byzantine agents, extreme values trimming is used in lines 9 and 18. 
\begin{algorithm}[h]
\caption{Hierarchical Byzantine-resilient Non-Bayesian Learning}
\label{alg: hierarchical: pairwise}
 \vskip 0.2\baselineskip
 {\normalsize
 \For{$j\in \cup_{i=1}^M \calV_i$}
 {
 \For{$\theta_1, \theta_2\in \Theta \text{such that }~ \theta_1\not=\theta_2$}
 {$r_0^j(\theta_1, \theta_2)\gets 0$\;}
} 

In parallel, for each hypothesis pair $\theta_1, \theta_2$ do: \\
\While{$t\ge 1$}{
 
 \eIf{Agent $j$ belongs to a network in $\calC$}
{ 
 Transmit $r_{t-1}^j(\theta_1, \theta_2)$ on all outgoing edges\;
 %
%


Filter the smallest and largest $F$ values, respectively, of the received log likelihood ratios $\tilde{r}_{t-1}^{j^{\prime}}(\theta_1, \theta_2)$ 

%
\vskip 0.2\baselineskip
$r_{t}^j(\theta_1, \theta_2) \gets \frac{\sum_{j^{\prime}\in \calI_j^*[t]} \tilde{r}_{t-1}^{j^{\prime}}(\theta_1, \theta_2) + r_{t-1}^j(\theta_1, \theta_2)}{|\calI_j^*[t]|+1} + \log \frac{\ell_j (s^j_{1, t} \mid \theta_1) }{\ell_j (s^j_{1, t} \mid \theta_2)}.$
}
{
\If{$t\mod \Gamma =0$}
{

\uIf{$M\ge 2F+1$}
{
The parameter server randomly chooses one representative from each of the $M$ networks 
and queries these representatives their local estimates\; 

}
\Else{
For each $i\in \calC$, randomly choose one agent in $\calV_i$ as network representative of iteration $t$. 
Choose $\pth{2F+1 - |\calC|}$ representatives from $\cup_{i\notin \calC}\calV_i$ uniformly at random as representatives. 

Queries these representatives their local estimates\;  
}

The parameter server removes messages with the largest $F$ and smallest $F$ values\; 

$\tilde{w}(t) \gets \frac{1}{|\tilde{\calR}(t)|}\sum_{j\in\tilde{\calR}(t)} m_{j}(t)$\; 

Broadcasts $\tilde{w}(t)$ to each of randomly chosen network representatives 
$j_1(t), \cdots, j_{\max\{2F+1, M\}}(t)$. 

\For{$\ell=1, \cdots, \max\{2F+1, M\}$}
{
\If{Agent $j_{\ell}(t)$ does not belong to a network in $\calC$}
{
$r_{t}^{j_{\ell}(t)}(\theta_1, \theta_2)  \gets \tilde{w}(t)$\; 
}
}
}

}
}
}
\end{algorithm}

\begin{assumption} 
\label{ass: sufficiency: Byzantine: non-Bayesian: mixing + information source}
There exist at least $F+1$ networks $G(\calV_i, \calE_i)$ each of which satisfies Assumptions \ref{ass: tight condition: single network} and \ref{ass: Byzantine: source identifiability}. 
\end{assumption}

\begin{theorem}
\label{thm: convergence: Byzantine: single network}
Suppose Assumption \ref{ass: sufficiency: Byzantine: non-Bayesian: mixing + information source} holds. 
For each normal agent, there exists a unique hypothesis $\tilde{\theta}$ such that $\forall \,\theta\not=\tilde{\theta}$: 
\[
\lim\sup_{t\diverge} r_{t}^j(\tilde{\theta}, \theta)\toas +\infty,  \text{and} \lim\inf_{t\diverge} r_{t}^j(\theta, \tilde{\theta})\toas -\infty. 
\]
\end{theorem}
\begin{remark}
Theorem \ref{thm: convergence: Byzantine: single network} is non-trivial. 
By \cite{su2019defending}, Assumption \ref{ass: sufficiency: Byzantine: non-Bayesian: mixing + information source} requires at least $F+1$ networks can reach consensus individually despite different learning rates. However, since the Byzantine agents can lie arbitrarily and the local signals are non-IID and noisy, agents in $\calC$ may not effectively propagate its local learning to agents in a different network. Particularly, in line 17, it is possible that the messages from the sample agents in $\calC$ are all filtered out by the PS. 
Though the pairwise linear dynamics are also considered in \cite{su2019defending}, formal analysis was missing and the sketched proof does not go through. This is because the KL divergence term shows up only when one of the hypothesis involved is the underlying truth $\theta^*$. 
\end{remark}

\begin{remark}
The Byzantine agents $\calA$ can be arbitrary subset of $\cup_{i=1}^M \calV_i$ as long as $\abth{\calA} \le F$. One interesting extreme case is when all the Byzantine agents are located in the same sub-network. Assumption \ref{ass: sufficiency: Byzantine: non-Bayesian: mixing + information source} implies that $F<\frac{1}{3} n_i$ for each $i\in \calC$.    

It is worth noting that for a sub-network outside $\calC$, even if the majority of the agents are Byzantine, our algorithm still enables the normal agents to learn $\theta^*$. 
\end{remark}



\newpage 

\bibliographystyle{abbrv}
\bibliography{reference_Learning,PSDA_DL}



\clearpage


\newpage 

\appendices

\section{Algorithm: Dropping link resilience}
\label{app: algorithm: dropping link}
The Algorithm in Section \ref{sec: algorithm: dropping link} is formally described in Algorithm \ref{alg:push-sum hierarchical FL: dropping link}. 
\begin{algorithm}
\caption{Non-Bayesian Learning: Dropping Link}
\label{alg:push-sum hierarchical FL: dropping link}
{\small {\em Initialization}: 
For each sub-network $i=1, \cdots, M$: $z_j^i[0]=\bm{0}\in R^d$, $m^i_j[0]=1\in R,$ 
$\sigma_j^i[0]={\bf 0}\in R^d$, $\tilde{\sigma}_j^i[0]=0\in R$, and $\rho_{j^{\prime}j}[0]={\bf 0}\in R^d$, $\tilde{\rho}_{j^{\prime}j}[0]=0\in R$ for each incoming link, i.e., $j^{\prime} \in \calI_j^i$.

\vskip 0.2\baselineskip 
In parallel, each client $j\in \calV_i$ does:\\
\For{$t\ge 1$}
{ 
$\sigma_j^{i+}[t]  \gets  \sigma_j^i[t-1] + \frac{z_j^i[t-1]}{d_j^i[t]+1}$,
$\tilde{\sigma}_j^{i+}[t] \gets \tilde{\sigma}_j^i[t-1] + \frac{m_j^i[t-1]}{d_j^i[t]+1}$\;

Broadcast $\pth{\sigma^{i+}_j[t], \tilde{\sigma}^{i+}_j[t]}$ to outgoing neighbors\;

\For {each incoming link $(j^{\prime},j)\in \calE_i$}
{\eIf{message $\pth{\sigma^{i+}_{j^{\prime}}[t], \tilde{\sigma}^{i+}_{j^{\prime}}[t]}$ is received}
{$\rho^i_{j^{\prime}j}[t] \gets \sigma^{i+}_{j^{\prime}}[t]$, ~~ $\tilde{\rho}^i_{j^{\prime}j}[t] \gets \tilde{\sigma}^{i+}_{j^{\prime}}[t]$\;}
{ $\rho^i_{j^{\prime}j}[t] \gets \rho^i_{j^{\prime}j}[t-1]$, ~~$\tilde{\rho}^i_{j^{\prime}j}[t] \gets \tilde{\rho}^i_{j^{\prime}j}[t-1]$\;}
$ z_j^{i+}[t] \gets \frac{z_j^{i}[t-1]}{d_j^i[t]+1} +  \sum_{j^{\prime}\in \calI_j^i} \pth{\rho_{j^{\prime}j}[t] - \rho_{j^{\prime}j}[t-1]}$, 
$m_j^{i+}[t]  \gets \frac{m_j^i[t-1]}{d_j^i[t]+1} + \sum_{j^{\prime}\in \calI_j^i} \pth{\tilde{\rho}_{j^{\prime}j}[t] -\tilde{\rho}_{j^{\prime}j}[t-1]}$.
}

$\sigma^i_j[t]  \gets  \sigma^{i+}_j[t] + \frac{z_j^{i+}[t]}{d_j^i[t]+1}$,
$\tilde{\sigma}^i_j[t]  \gets  \tilde{\sigma}^{i+}_j[t] + \frac{m_j^{i+}[t]}{d_j^i[t]+1}$,
$z_j^i[t]  \gets \frac{z_j^{i+}[t]}{d_j^i[t]+1}$,
$m_j^i[t] \gets \frac{m_j^{i+}[t]}{d_j^i[t]+1}$\; 

Obtain measurement $s_j^i(t)$\; 
\For{$\ell=1, \cdots, m$}
{
$z_j^i(\theta_{\ell}, t) \gets  z_j^i(\theta_{\ell}, t) + \log\pth{\ell(s_j^i(t)\mid \theta_{\ell})}$\;
}
$\mu_j^i(\cdot, t) \gets \prod_{\mu\in \Delta_{\Theta}}^{\varphi}\pth{\frac{z_j^i(\cdot, t)}{m_j^i(t)}, 1}$\; 
}

\If{$j$ is a designated agent of network $S_i$}
{
\If{$t\mod \Gamma =0$}
{
Send $\frac{1}{2}z_j^i[t]$ and $\frac{1}{2}m_j^i[t]$ to the PS\; 

Upon receiving messages from the PS {\bf do} \\
update  
$z_j^{i}[t]\gets \frac{1}{2}z_j^{i}[t] + \frac{1}{2M}\sum_{i=1}^M z_{i_0}^i[t]$\; 
$m_j^{i}[t]\gets \frac{1}{2}m_j^{i}[t] + \frac{1}{2M}\sum_{i=1}^M m_{i_0}^i[t]$\;
}
} 

\If{$t\mod \Gamma =0$}
{
The PS does the following: 

Wait to receive $z_{i_0}^i[t]$ and $m_{i_0}^i[t]$ from each designated agent of the $M$ networks\; 

Compute and send $\frac{1}{M}\sum_{i=1}^M \frac{1}{2}z_{i_0}^i[t]$ and $\frac{1}{M}\sum_{i=1}^M \frac{1}{2}m_{i_0}^i[t]$ to all designated agents $i_0$ for $i=1, \cdots, M$. 
} 

}
\end{algorithm}
For agent $j$ in network $i$, the variable $z_{j}^i(\cdot, t)\in R^m$ as its local average of the log likelihood for each $\ell=1, \cdots, m$ and $m_j^i(t)\in R$ as its local mass.  
For ease of exposition, let 
\begin{align}
\label{eq: log likelihood}
L_j^i (\theta_{\ell},t) = \log\pth{\ell(s_j^i(t)\mid \theta_{\ell})}.  
\end{align}
%
Adapting the notation from \cite{UnreliableConvergence}, we have for each agent $j$ in network $i$,
\begin{equation}
\label{eq: dropping link: matrix iterative: local}
\begin{aligned}
z_j^i(\theta_{\ell}, t) &= \sum_{r=0}^{t-1} \sum_{j^{\prime}=1}^{n_i} L_{j^{\prime}}^i (\theta_{\ell},r)\bm{\Psi}_{j^{\prime}j}(r,t), \\
m_j^{i}[t] &= \sum_{j^{\prime}=1}^{n_i} \bm{\Psi}_{j^{\prime}j}(1,t),
\end{aligned}    
\end{equation}
where $\bm{\Psi}_{j^{\prime}j}(r,t) = \bm{M}[t]\cdots \bm{M}[r]$ for $r\le t$. 

Intuitively, the variable $z_j^i(\theta_{\ell}, t)$ stores the {\em locally} averaged log-likelihood with respect to hypothesis $\theta_{\ell}$ at iteration $t$. 
Define $\bar{z}(\theta_{\ell})$ to be the {\em globally} averaged log-likelihood with respect to hypothesis $\theta_{\ell}$ at iteration $t$, i.e., 
\begin{align}
\label{eq: eq: dropping link: matrix iterative: global}
\bar{z}(\theta_{\ell}) = \frac{1}{N}\sum_{i=1}^M \sum_{j=1}^{n_i} z_j^i(\theta_{\ell}, t) 
= \frac{1}{N}\sum_{i=1}^M \sum_{j=1}^{n_i} \sum_{r=1}^t L_{j^{\prime}}^{i^{\prime}}(\theta_{\ell}, r). 
\end{align}

\subsection{Proof of Theorem \ref{thm: iterative}}
Without loss of generality, let's assume $\theta_1 = \theta^*$. If this is not true, then we can permute the ordering of $\theta_1, \cdots, \theta_m$ so that under the permuted ordering, it is true that $\theta_1 = \theta^*$. It is worth noting that the permutation is only used for analysis purpose. The execution of Algorithm \ref{alg:push-sum hierarchical FL} does not rely on the knowledge of the permutation. 

We characterize the dynamics of $\log \frac{\mu_j^i(\theta_{\ell}, t)}{\mu_j^i(\theta_1, t)}$ for $\ell=2, \cdots, m$.  
\begin{align*}
\log \frac{\mu_j^i(\theta_{\ell}, t)}{\mu_j^i(\theta_1, t)} 
& \overset{(a)}{=}  \log \pth{\frac{\exp\pth{\frac{z_j^i(\theta_{
\ell}, t)}{m_j^i(t)}}}{\exp\pth{\frac{z_j^i(\theta_{1}, t)}{m_j^i(t)}}} } \\
& = \frac{z_j^i(\theta_{
\ell}, t)-z_j^i(\theta_{1}, t)}{m_j^i(t)} \\
& = \underbrace{\frac{z_j^i(\theta_{
\ell}, t)}{m_j^i(t)} - \bar{z}(\theta_{\ell}, t)}_{(A)} +  \underbrace{\bar{z}(\theta_{1}, t) - \frac{z_j^i(\theta_{1}, t)}{m_j^i(t)}}_{(B)} \\
& \qquad + \underbrace{\bar{z}(\theta_{\ell}, t) - \bar{z}(\theta_{1}, t) }_{(C)},
\end{align*} 
where equality (a) holds because that $\mu_0(\theta_{
\ell}) = \frac{1}{m}$ $\forall\, \theta_{\ell}\in \Theta$. 

\vskip \baselineskip

\noindent{\bf Bounding (C).}  
We first bound (C) as follows. 
\begin{align*}
&\bar{z}(\theta_{\ell}, t) - \bar{z}(\theta_{1}, t)\\ 
&= \frac{1}{N}\sum_{i=1}^M \sum_{j=1}^{n_i} \sum_{r=1}^t L_{j^{\prime}}^{i^{\prime}}(\theta_{\ell}, r) 
- \frac{1}{N}\sum_{i=1}^M \sum_{j=1}^{n_i} \sum_{r=1}^t L_{j^{\prime}}^{i^{\prime}}(\theta_1, r) \\
& = \frac{1}{N}\sum_{i=1}^M \sum_{j=1}^{n_i} \sum_{r=1}^t \log\frac{\ell(s_j^i(t)\mid \theta_{\ell})}{\ell(s_j^i(t)\mid \theta_{1})}.  
\end{align*}
Recall that $\sup_{w_{i_j}\in \calS_{i_j}, \text{and }\theta, \theta^{\prime}\in \Theta} \log\frac{\ell_{i_j}(w_{i_j} | \theta)}{\ell_{i_j}(w_{i_j} | \theta^{\prime}}\le L$ for some positive constant $L>0$.  
Since $\theta_1 = \theta^*$, i.e., the signals $s_j$ is generated according to distribution $\ell(\cdot \mid \theta_1)$, 
by Hoeffding's inequality, we have that for any given $t$, with probability at least $1-\delta$ for some given accuracy requirement $\delta>0$ 
\begin{align*}
 &\frac{1}{N}\sum_{i=1}^M \sum_{j=1}^{n_i} \sum_{r=1}^t \log\frac{\ell(s_j^i(t)\mid \theta_{\ell})}{\ell(s_j^i(t)\mid \theta_{1})} \\
 &\le \expect{\frac{1}{N}\sum_{i=1}^M \sum_{j=1}^{n_i} \sum_{r=1}^t \log\frac{\ell(s_j^i(t)\mid \theta_{\ell})}{\ell(s_j^i(t)\mid \theta_{1})}} + tL\sqrt{\frac{2}{t}\log \frac{1}{\delta}}\\
 & =  -\frac{t}{N} D_{KL}\pth{\bm{\ell}(\cdot \mid \theta_{1}) \| ~\bm{\ell}(\cdot \mid \theta_{\ell})} + L\sqrt{2t\log \frac{1}{\delta}}. 
\end{align*}
For ease of exposition, define 
\begin{align*}
D_{\text{KL}}\pth{\theta_1, \theta_{\ell}} ~ = ~ D_{KL}\pth{\bm{\ell}(\cdot \mid \theta_{1}) \| ~\bm{\ell}(\cdot \mid \theta_{\ell})}. 
\end{align*}

\vskip \baselineskip

\noindent The two terms (A) and (B) can be bounded similarly. Henceforth, we focus on bounding term (B). The analysis is also  analogously to our analysis for the state estimation problem.  

\noindent{\em Bounding (B).}
\begin{align*}
&\bar{z}(\theta_{1}, t) - \frac{z_j^i(\theta_{1}, t)}{m_j^i(t)}\\
& = \frac{1}{N}\sum_{i^{\prime}=1}^M \sum_{j^{\prime}=1}^{n_i} \sum_{r=1}^t L_{j^{\prime}}^{i^{\prime}}(\theta_1, r) \\
& \qquad -  \frac{\sum_{r=1}^{t} \sum_{i^{\prime}=1}^M \sum_{j^{\prime}=1}^{n_i}L_{j^{\prime}}^{i^{\prime}}(\theta_1, r) \bm{\Psi}_{j^{\prime},j}(r,t)}{\sum_{i=1}^M\sum_{j^{\prime}=1}^{n_i} \bm{\Psi}_{j^{\prime}, j}(1,t)}
 \\
& = \frac{\sum_{r=1}^{t} \sum_{i=1}^M\sum_{j^{\prime}=1}^{n_i} L_{j^{\prime}}^{i^{\prime}}(\theta_1, r)  \sum_{k=1}^{N} \pth{ {\bf \Psi}_{k,j}(1, t) -  {\bf \Psi}_{j^{\prime},j}(r, t)}   }{ N \sum_{i=1}^M\sum_{j^{\prime}=1}^{n_i} {\bf \Psi}_{j^{\prime},j}(1,t)}.
\end{align*}
Thus, 
\begin{align*}
&\norm{\bar{z}(\theta_{1}, t) - \frac{z_j^i(\theta_{1}, t)}{m_j^i(t)}}\\ 
& \le \frac{L\sum_{r=1}^{t} \sum_{i=1}^M\sum_{j^{\prime}=1}^{n_i} \abth{\sum_{k=1}^{N} \pth{ {\bf \Psi}_{k,j}(1, t) -  {\bf \Psi}_{j^{\prime},j}(r, t)}}}{ N \sum_{i=1}^M\sum_{j^{\prime}=1}^{n_i} {\bf \Psi}_{j^{\prime},j}(1,t)}\\
& \le \frac{ L \sum_{r=1}^{t} \sum_{i=1}^M\sum_{j^{\prime}=1}^{n_i} \abth{ {\bf \Psi}_{k,j}(1, t) -  {\bf \Psi}_{j^{\prime},j}(r, t)}}{{N}^2   \frac{1}{4M^2}\pth{\min_{i\in [M]}\beta_i}^{2D^*B}}\\
& = \frac{ 4M^2L \sum_{r=1}^{t} \sum_{i=1}^M\sum_{j^{\prime}=1}^{n_i} \abth{ {\bf \Psi}_{k,j}(1, t) -  {\bf \Psi}_{j^{\prime},j}(r, t)}}{{N}^2 \pth{\min_{i\in [M]}\beta_i}^{2D^*B}},
\end{align*}
where the last inequality follows from Lemma \ref{lm: entire matrix lower bound}. 
In addition, 
\begin{align*}
&\abth{{\bf \Psi}_{k,j}(1, t) -  {\bf \Psi}_{j^{\prime},j}(r, t)}\\
& = \abth{ \sum_{p=1}^N \bm{\Psi}_{k,p}(1, r-1) {\bf \Psi}_{p,j}(r, t) -  {\bf \Psi}_{j^{\prime},j}(r, t)} \\
& \le \sum_{p=1}^N \bm{\Psi}_{k,p}(1, r-1) \abth{{\bf \Psi}_{p,j}(r, t) -  {\bf \Psi}_{j^{\prime},j}(r, t)}\\
& \le \sum_{p=1}^N \bm{\Psi}_{k,p}(1, r-1) \gamma^{\frac{t-r}{2\Gamma}}\\
& = \gamma^{\frac{t-r+1}{2\Gamma}}. 
\end{align*}
where the last inequality follows from the key intermediate results in bounding the convergence rate of the matrix product $\bm{\Phi}$, which is omitted due to lack of space. 
So, term (B) can be bounded as 
\begin{multline}
\label{eq: dlink: bound (B)} 
\norm{\bar{z}(\theta_{1}, t) - \frac{z_j^i(\theta_{1}, t)}{m_j^i(t)}} 
\le \frac{ 4M^2L \sum_{r=1}^{t}\gamma^{\frac{t-r+1}{2\Gamma}}}{N \pth{\min_{i\in [M]}\beta_i}^{2D^*B}} \\
\le \frac{ 4M^2L \gamma^{\frac{1}{2\Gamma}}}{N \pth{1-\gamma^{\frac{1}{2\Gamma}}}\pth{\min_{i\in [M]}\beta_i}^{2D^*B}}. 
\end{multline}

Therefore, we conclude for any given $\delta\in (0,1)$, the following holds with probability at least $1-\delta$:  For all $\theta_{\ell}\in \Theta$ for $\ell=2, \cdots, m$,   
\begin{multline*}
\log \frac{\mu_j^i(\theta_{\ell}, t)}{\mu_j^i(\theta_1, t)}
\le \frac{ 8M^2L \gamma^{\frac{1}{2\Gamma}}}{N \pth{1-\gamma^{\frac{1}{2\Gamma}}}\pth{\min_{i\in [M]}\beta_i}^{2D^*B}}  \\
-\frac{t}{N} D_{KL}\pth{\theta_{1}\|  \theta_{\ell}} + L\sqrt{2t\log \frac{m}{\delta}},
\end{multline*} 
which goes to $-\infty$ as $t\diverge$. To see this, for any given $\delta\in (0,1)$, as long as 
$t\ge \frac{8N^2L^2\log\frac{m}{\delta}}{D^2_{KL}\pth{\theta_1\|\theta_{\ell}}}$, it holds that 
\begin{multline*}
 -\frac{t}{N} D_{KL}\pth{\theta_{1}\|  \theta_{\ell}} + L\sqrt{2t\log \frac{m}{\delta}} \\
 \le -\frac{t}{N} D_{KL}\pth{\theta_{1}\|  \theta_{\ell}} + \frac{1}{2}\frac{t}{N}D_{KL}\pth{\theta_{1}\|  \theta_{\ell}}\\
 = - \frac{1}{2}\frac{t}{N}D_{KL}\pth{\theta_{1}\|  \theta_{\ell}}. 
\end{multline*}

\section{Byzantine resilience}
\label{app: Byzantine}

\subsection{Proof of Theorem \ref{thm: convergence: Byzantine: single network} } 
\label{subsec: Byzantine consensus: convergence analysis}
We first show Theorem \ref{thm: convergence: Byzantine: single network} for each normal agent that belongs to a network in $\calC$. 
Define 
\begin{align}
\label{eq: minimal soucre drift}
D_{KL}^* : =\min_{\theta\in  \Theta\setminus \{\theta^*\}} \min_{\calH \in \calG_{\text{info}}} \sum_{j\in \calS_{\calH}}D_{KL}\pth{\ell_j(\cdot |\theta^*)\parallel\ell_j(\cdot |\theta)}.       
\end{align}
\subsubsection{Convergence in a network in $\calC$}
\label{subsubsec: convergence in C}
\begin{lemma}
\label{lm: pairwise learning}
Fix any network $i$ in $\calC$. 
Let $n_i = \abth{\calV_i}$ and $\phi_i = \abth{\calV_i \setminus \calA}$. 
Let $j\in \calV_i\setminus \calA$ be an arbitrary non-Byzantine agent. 
For any $\theta \not=\theta^*$, the following holds:
\begin{align*}
\lim_{t\diverge}\frac{1}{t^2}r_{t}^j(\theta^*, \theta)\ge  \frac{1}{2}{\beta^{\chi_i(n_i-\phi_i)}} D_{KL}^*, ~\text{and} ~ \\
\lim_{t\diverge}\frac{1}{t^2} r_{t}^j(\theta, \theta^*)\le  -\frac{1}{2}{\beta^{\chi_i(n_i-\phi_i)}} D_{KL}^* ~~~~ \text{almost surely},
\end{align*}
where $\beta \triangleq \min_{i\in \calC}\min_{j\in \calV_i\setminus \calA}\frac{1}{2(d^i_j-2F)+1}, $ 
recalling that $d_j^i$ is the incoming degree of agent $j$ which belongs to $S_i$. 

\end{lemma}
\vskip \baselineskip
\noindent{\em Proof.}
Since the intersection of finitely many almost surely events is also almost surely, i.e., if $\prob{D_k} =1$ for $k=1, \cdots, n$, then 
\begin{align*}
\prob{\cap_{k=1}^n D_k} =1,      
\end{align*}
It is enough to consider the convergence for each pair of $\theta_1$ and $\theta_2$ separately. 

    By  \cite{vaidya2012matrix}, we know that for each pair of hypotheses $\theta_1$ and $\theta_2$, there exists a row-stochastic matrix ${\bf M}^{1, 2}[t]\in R^{(n_i-\phi_i)\times (n_i-\phi_i)}$ such that
\begin{align}
\label{update pairwise}
r_{t}^j(\theta_1, \theta_2)= \sum_{j^{\prime}=1}^{n_i-\phi_i} {\bf M}^{1,2}_{jj^{\prime}}[t] r_{t-1}^j(\theta_1, \theta_2)+  \log \frac{\ell_j (s^j_{1, t} \mid \theta_1) }{\ell_j (s^j_{1, t} \mid \theta_2)}.
\end{align}
It is worth noting that the above matrix $\bm{M}$ is different from that for the dropping-link setup. Here $\bm{M}\in R^{(n_i-\phi_i)\times (n_i-\phi_i)}$ is defined for each network $i\in \calC$, where in the dropping-link setup, $\bm{M}\in R^{N\times N}$ is defined for the entire hierarchical system. 
In addition,  matrix ${\bf M}^{1,2}[t]$ depends on the choice of hypotheses $\theta_1$ and $\theta_2$, 
and is time-varying. 
The reason of that ${\bf M}^{1,2}[t]$ is time-varying is two-fold: 
(1) The log likelihood ratio of the cumulative signals $\log \frac{\ell_j (s^j_{1, t} \mid \theta_1) }{\ell_j (s^j_{1, t} \mid \theta_2)}$ is changing over time due to the obtain of new signal and the randomness in the signal; and 
(2) the Byzantine agents can adaptively calibrate their malicious messages based on algorithm execution up to time $t$.  

For a given pair of hypotheses $\theta_1$ and $\theta_2$, let ${\bf r}_{t}(\theta_1, \theta_2)\in R^{n_i-\phi_i}$ be the vector that stacks $r_{t}^j(\theta_1, \theta_2)$. The evolution of ${\bf r}(\theta_1, \theta_2)$ can be compactly written as
\begin{align}
\label{update pairwise vector}
\nonumber
{\bf r}_{t}(\theta_1, \theta_2)&=  {\bf M}^{1,2}[t] {\bf r}_{t-1}(\theta_1, \theta_2)+  \sum_{r=1}^t \calL_r(\theta_1, \theta_2) \\
&=  \sum_{r=1}^t {\bf \Phi}^{1,2}(t, r+1) \sum_{k=1}^r \calL_k(\theta_1, \theta_2),
\end{align}
where ${\bf \Phi}^{1,2}(t, r+1)\triangleq {\bf M}^{1, 2}[t] {\bf M}^{1, 2}[t-1] \cdots {\bf M}^{1, 2}[r+1]$ for $r\le t$, ${\bf \Phi}^{1,2}(t, t)\triangleq {\bf M}^{1, 2}[t]$ and ${\bf \Phi}^{1,2}(t, t+1)\triangleq {\bf I}$.  
The last equality holds because $\mu_j(\theta_{\ell}, 0) = \frac{1}{m}$, which immediately leads to ${\bf r}_{t-1}(\theta_1, \theta_2) = \bm{0}$.

Using coefficients of ergodicity \cite{Hajnal58}, under Assumption \ref{ass: tight condition: single network},  it has been shown \cite{vaidya2014iterative} that
\begin{align}
\label{mixing}
\lim_{t\ge r,~ t\diverge}{\bf \Phi}^{1,2}(t, r)=\ones {\bf \pi}^{1,2}(r),
\end{align}
where ${\bf \pi}(r)^{1,2}\in R^{(n_i-\phi_i)}$ is a row stochastic vector, and $\ones$ is the column vector with each entry being $1$.

Moreover, by the proof of \cite[Lemma 4]{vaidya2014iterative}, we know that:  
For any $r\ge 1$, there exists a reduced graph $\calH [r]$ with source component $\calS\pth{\calH[r]}$ such that $\pi^{1,2}_i(r)\ge \beta^{\chi_i(n_i-\phi_i)}$ for each $j\in \calS_r$. 

To prove $\lim_{t\diverge}\frac{1}{t^2}r_{t}^j(\theta^*, \theta)\ge  \frac{1}{2}{\beta^{\chi_i(n_i-\phi_i)}} D_{KL}^*$, without loss of generality,  let $\theta_1 = \theta^*$. Clearly, $\theta_2\not=\theta_1=\theta^*$. 
\begin{multline*}
{\bf r}_{t}(\theta_1, \theta_2) =  \sum_{r=1}^t {\bf \Phi}^{1,2}(t, r+1) \sum_{k=1}^r \calL_k(\theta_1, \theta_2) \\
= \sum_{r=1}^t \left({\bf \Phi}^{1,2}(t, r+1) \sum_{k=1}^r \calL_k(\theta_1, \theta_2)  \right.\\
- \left.r\bm{1} \sum_{j^{\prime}=1}^{n_i-\phi_i} \pi_{j^{\prime}}(r+1) D_{KL}^{j^{\prime}}(\theta_1, \theta_2) \right.\\
+ \left.r\bm{1} \sum_{j^{\prime}=1}^{n_i-\phi_i} \pi_{j^{\prime}}(r+1) D_{KL}^{j^{\prime}}(\theta_1, \theta_2)\right). 
\end{multline*}
For each $j\in \calV_i\setminus \calA$, we have  
\begin{align*}
r_t^j\pth{\theta_1, \theta_2} = 
\sum_{r=1}^t  \left(\sum_{j^{\prime}=1}^{n_i-\phi_i}{\bf \Phi}_{jj^{\prime}}^{1,2}(t, r+1) \sum_{k=1}^r \calL_k^{j^{\prime}}(\theta_1, \theta_2) \right. \\
\underbrace{
- \left.r \sum_{j^{\prime}=1}^{n_i-\phi_i} \pi_{j^{\prime}}(r+1) D_{KL}^{j^{\prime}}(\theta_1, \theta_2) \right)}_{(A)} \\
\qquad + \underbrace{\sum_{r=1}^t r \sum_{j^{\prime}=1}^{n_i-\phi_i} \pi_{j^{\prime}}(r+1) D_{KL}^{j^{\prime}}(\theta_1, \theta_2)}_{(B)}.  
\end{align*}

We bound $(B)$ first. We have 
\begin{align*}
&\sum_{r=1}^t r \sum_{j^{\prime}=1}^{n_i-\phi_i} \pi_{j^{\prime}}(r+1) D_{KL}^{j^{\prime}}(\theta_1, \theta_2)   \\
& \ge \sum_{r=1}^t r \sum_{j^{\prime}\in \calS_r} \pi_{j^{\prime}}(r+1) D_{KL}^{j^{\prime}}(\theta_1, \theta_2) \\
& \ge \sum_{r=1}^t r  {\beta^{\chi_i(n_i-\phi_i)}} \sum_{j^{\prime}\in \calS_r} D_{KL}^{j^{\prime}}(\theta_1, \theta_2) \\
& \ge \sum_{r=1}^t r  {\beta^{\chi_i(n_i-\phi_i)}} D_{KL}^* \\
& = \frac{t(t+1)}{2}  {\beta^{\chi_i(n_i-\phi_i)}} D_{KL}^*.  
\end{align*}
Thus, let $t\diverge$, it holds that 
\begin{align*}
  \lim_{t\diverge} \sum_{r=1}^t r \sum_{j^{\prime}=1}^{n_i-\phi_i} \pi_{j^{\prime}}(r+1) D_{KL}^{j^{\prime}}(\theta_1, \theta_2) ~~\to ~~  +\infty.   
\end{align*}
Specifically, 
\begin{align*}
& \lim_{t\diverge} \frac{1}{t^2} \sum_{r=1}^t r \sum_{j^{\prime}=1}^{n_i-\phi_i} \pi_{j^{\prime}}(r+1) D_{KL}^{j^{\prime}}(\theta_1, \theta_2) \\
&\ge  \frac{1}{2}{\beta^{\chi_i(n_i-\phi_i)}} D_{KL}^*. 
\end{align*}

To bound $(A)$, we first note that when $\theta_1= \theta^*$, it holds that 
\begin{align*}
\mathbb{E}_{s\sim \ell(\cdot \mid \theta^*)} \qth{\calL_k^{j^{\prime}}(\theta^*, \theta_2)} 
&= \mathbb{E}_{s\sim \ell(\cdot \mid \theta^*)} \qth{\log \frac{\ell_{j^{\prime}}\pth{s_t^{j^{\prime}}\mid \theta^*}}{\ell_{j^{\prime}}\pth{s_t^{j^{\prime}}\mid \theta_2}}} \\
&= D_{KL}^{j^{\prime}}\pth{\theta^*\| \theta_2}. 
\end{align*}
Thus, following the same argument as the proof of \cite[Lemma 3]{su2019defending}, we can show that 
\begin{multline*}
\frac{1}{t^2} \sum_{r=1}^t \left( r \sum_{j^{\prime}=1}^{n_i-\phi_i} \pi_{j^{\prime}}(r+1) \calL_k^{j^{\prime}}(\theta_1, \theta_2) - \right.\\
\left. r \sum_{j^{\prime}=1}^{n_i-\phi_i} \pi_{j^{\prime}}(r+1) D_{KL}^{j^{\prime}}(\theta_1, \theta_2)\right) ~~ \overset{a.\,s.\,}{\longrightarrow} ~~ 0.  
\end{multline*}
Hence, for any $\theta\not=\theta^*$, it holds that 
\begin{align*}
  \lim_{t\diverge}\frac{1}{t^2} r_t^j(\theta^*, \theta) ~ \ge ~ \frac{1}{2}\beta_m^{\chi_i(n_i-\phi_i)} D_{KL}^*, ~~~ \text{almost surely.}    
\end{align*} 
\vskip \baselineskip 
It remains to prove 
\[
\lim_{t\diverge}\frac{1}{t^2}r_{t}^j(\theta, \theta^*) \le  -\frac{1}{2}\beta_m^{\chi_i(n_i-\phi_i)} D_{KL}^*, ~~~ \text{almost surely.} 
\]
It is sufficient to consider the scenario when $\theta_2 = \theta^*$. 
We have 
\begin{align*}
{\bf r}_{t}(\theta_1, \theta^*) &=  \sum_{r=1}^t {\bf \Phi}^{1,2}(t, r+1) \sum_{k=1}^r \calL_k(\theta_1, \theta^*) \\
& = \sum_{r=1}^t \left({\bf \Phi}^{1,2}(t, r+1) \sum_{k=1}^r \calL_k(\theta_1, \theta^*) \right.\\
&+ \left.r\bm{1} \sum_{j^{\prime}=1}^{n_i-\phi_i} \pi_{j^{\prime}}(r+1) D_{KL}^{j^{\prime}}(\theta^*\|\theta_1) \right.\\
&- \left. r\bm{1} \sum_{j^{\prime}=1}^{n_i-\phi_i} \pi_{j^{\prime}}(r+1) D_{KL}^{j^{\prime}}(\theta^*\|\theta_1)\right).   
\end{align*}
For each $j\in \calV_i\setminus \calA$, we have 
\begin{align*}
r_t^j\pth{\theta_1, \theta^*} &= \sum_{r=1}^t \left(\sum_{j^{\prime}=1}^{n_i-\phi_i}{\bf \Phi}_{jj^{\prime}}^{1,2}(t, r+1) \sum_{k=1}^r \calL_k^{j^{\prime}}(\theta_1, \theta^*) \right.\\
& \underbrace{\left.+ r \sum_{j^{\prime}=1}^{n_i-\phi_i} \pi_{j^{\prime}}(r+1) D_{KL}^{j^{\prime}}(\theta^*\|\theta_1) \right)}_{(C)}   \\
&  \underbrace{-\sum_{r=1}^t r \sum_{j^{\prime}=1}^{n_i-\phi_i} \pi_{j^{\prime}}(r+1) D_{KL}^{j^{\prime}}(\theta^*\|\theta_1)}_{(D)}.  
\end{align*}
Similar as before, we can show that 
\begin{align*}
&  \lim_{t\diverge}\frac{1}{t^2}\sum_{r=1}^t r \sum_{j^{\prime}=1}^{n_i-\phi_i} \pi_{j^{\prime}}(r+1) D_{KL}^{j^{\prime}} (\theta^*\|\theta_1)  \\
&\ge \frac{1}{2}{\beta^{\chi_i(n_i-\phi_i)}} D_{KL}^*. 
\end{align*}
Thus, 
\begin{align*}
&- \lim_{t\diverge}\frac{1}{t^2}\sum_{r=1}^t r \sum_{j^{\prime}=1}^{n_i-\phi_i} \pi_{j^{\prime}}(r+1) D_{KL}^{j^{\prime}} (\theta^*\|\theta_1)\\
&\le = \frac{1}{2}{\beta^{\chi_i(n_i-\phi_i)}} D_{KL}^*.    
\end{align*}
 
In addition, since 
\begin{align*}
\calL_k^{j^{\prime}}(\theta_1, \theta^*) = - \calL_k^{j^{\prime}}(\theta^*, \theta_1),         
\end{align*}
by \cite[Lemma 3]{su2019defending}, we have 
\begin{align*}
&\lim_{t\diverge}\frac{1}{t^2} \sum_{r=1}^t  \left(\sum_{j^{\prime}=1}^{n_i-\phi_i}{\bf \Phi}_{jj^{\prime}}^{1,2}(t, r+1) \sum_{k=1}^r \calL_k^{j^{\prime}}(\theta_1, \theta^*) \right.\\
&\qquad \left.+ r \sum_{j^{\prime}=1}^{n_i-\phi_i} \pi_{j^{\prime}}(r+1) D_{KL}^{j^{\prime}}(\theta^*\|\theta_1)\right) ~ \overset{a.\,s.\,}{\longrightarrow} ~ 0.     
\end{align*}
Thus, we have 
\begin{align*}
\lim_{t\diverge}\frac{1}{t^2} r_{t}^j(\theta, \theta^*)\le  -\frac{1}{2}{\beta^{\chi_i(n_i-\phi_i)}} D_{KL}^* ~~ \text{almost surely.}     
\end{align*}

\hfill $\square$

The following corollary is an immediate consequence of Lemma \ref{lm: pairwise learning} by definition of convergence. 
\begin{corollary}
\label{cor: Byzantine: infinity}
For each agent $j$ in a network $i\in \calC$, 
\begin{align*}
\lim_{t\diverge} r_{t}^j(\theta^*, \theta)\toas +\infty, ~\text{and }~~ \lim_{t\diverge}r_{t}^j(\theta, \theta^*)\toas -\infty.
\end{align*}   
\end{corollary}

\begin{lemma}\cite[Proposition 1]{su2019defending}
\label{lm: uniqueness}
Fix a network in $i\in \calC$. 
Suppose Assumption \ref{ass: tight condition: single network} holds on graph $G\pth{\calV_i, \calE_i}$. 
Suppose there exists $\tilde{\theta}\in \Theta$ such that for any $\theta \not=\tilde{\theta}$, it holds that $\lim_{t\diverge} r_{t}^j(\tilde{\theta}, \theta)\toas +\infty$, and $\lim_{t\diverge} r_{t}^j(\theta, \tilde{\theta})\toas -\infty$. Then $\tilde{\theta}=\theta^*.$
\end{lemma}
\noindent{\em Proof.}
We prove this proposition by contradiction. Suppose there exists $\tilde{\theta}\not=\theta^*\in \Theta$ such that for any $\theta \not=\tilde{\theta}$, it holds that $\lim_{t\diverge} r_{t}^j(\tilde{\theta}, \theta)\toas +\infty$, and $\lim_{t\diverge} r_{t}^j(\theta, \tilde{\theta})\toas -\infty$. Then we know that $\lim_{t\diverge} r_{t}^j(\tilde{\theta}, \theta^*)\toas +\infty$ and $\lim_{t\diverge} r_{t}^j(\theta^*, \tilde{\theta})\toas -\infty$, contradicting Corollary \ref{cor: Byzantine: infinity}. 

\hfill $\square$

\subsubsection{Convergence at a general agent}
\label{sub: Byzantine convergence: general agent}
It remains to show the case when agent $j$ does not belong to any network in $\calC$.

\begin{theorem}
\label{thm: convergence: Byzantine: general agent}
For any non-Byzantine agent $j$ such that it does not belong to any of the networks in $\calC$. 
For all $\theta\not=\theta^*$, 
\[
\limsup_{t\diverge} \, r_{t}^j(\theta^*, \theta)\toas +\infty, ~~ \text{and} ~~  \liminf_{t\diverge} \, r_{t}^j(\theta, \theta^*)\toas -\infty. 
\]
\end{theorem}

\noindent{\em Proof.}
We focus on the scenario where $M\ge 2F+1$. The analysis can be easily adapted for the scenario where $M\le 2F$. 
Without loss of generality, let $S_1$ be the network that agent $j$ belongs to, i.e., $j\in \calV_1\setminus \calA$. 

Under Algorithm \ref{alg: hierarchical: pairwise}, for each $t$ such that $t\mod \Gamma =0$, agent $j$ is selected as the representative of network $S_1$ with probability $\frac{1}{n_1}>0$. 
Let $j_i\in \calV_i\setminus \calA$ be an arbitrary non-Byzantine agent for $i=2, \cdots, M$.   
Formally, we define a sequence of events as follows: For $k = 1, 2, \cdots $
\begin{align}
\label{eq: representative events}
A_k : = \{\omega: j_1(k\Gamma) = j, ~ \text{and}~ j_i(k\Gamma) = j_i ~ \forall ~ i\not=1\}. 
\end{align}
Let $p_k = \prob{A_k}$ for $k=1, 2, \cdots$. It is easy to see that $p_k = \frac{1}{n_1}\prod_{i=2}^{M}\frac{1}{n_i} = \prod_{i=1}^{M}\frac{1}{n_i}$ . 
Since $\sum_{k=1}^{\infty} p_k = \sum_{k=1}^{\infty}  \prod_{i=1}^{M}\frac{1}{n_i} = \infty$, and $A_1, A_2, \cdots$ are mutually independent, by Borel-Cantelli lemma \cite[Lemma 1.3]{hajek2015random}, we know 
\begin{align}
 \label{eq: gossiping infinitely often}   
 \mathbb{P}\sth{A_k ~\text{infinitely often}}=1, 
\end{align}
where $A_k \text{infinitely often} = \cap_{n\ge 1} \pth{\cup_{k\ge n} A_k}. $
That is, with probability 1 (almost surely), agents $j$ and $j_i$ for $i\in \calC$ are selected infinitely many times. 
Let $\tau_1, \tau_2, \cdots$ be the time indices at which agent $j$ is selected. 

Let $\omega$ be a sample path in which each of the network in $\calC$ learn $\theta^*$ independently, and that agent $j$ is selected as the representative of network $S_1$ infinitely often. 
Let $t^*$ be the time index such that for all $t\ge t^*$, $r_t^{j^{\prime}} \ge \frac{1}{2}{\beta^{\chi_i(n_i-\phi_i)}} D_{KL}^* t^2$ for all $i\in \calC$ and $j^{\prime}\in \calV_i\setminus \calA$. 
By Theorem \ref{thm: convergence: Byzantine: single network} and Eq.\eqref{eq: gossiping infinitely often}, we know that 
\[
\prob{\text{all such }\omega} = 1. 
\]
Notably, $t^*$ may change as the sample path $\omega$ changes. 
Henceforth, we fix one such sample path. 
Let 
\begin{align*}
c_{\min} &: = \min_{i\in \calC, j^{\prime}\in \calV_i\setminus \calA, ~ t<t^*} r_t^{j^{\prime}}(\theta_1, \theta_2), \\
c_{\max} : &= \min_{i\in \calC, j^{\prime}\in \calV_i\setminus \calA, ~ t<t^*} r_t^{j^{\prime}}(\theta_1, \theta_2), ~~ \text{where}~  \theta_1 = \theta^*.
\end{align*}

By definition, $\tilde{w}(t) = \frac{1}{|\tilde{\calR}(t)|}\sum_{j\in\tilde{\calR}(t)} m_{j}(t)$. 
For any $\tau_{r}$,  none of the representatives are Byzantine.   
Hence, we are able to rewrite (via two steps) $\tilde{w}(\tau_{r})$ in a form in which at least $M-F$ representatives have non-trivial influence on $j$. 
Specifically, 
\begin{align*}
\tilde{w}(\tau_{r}) = \frac{1}{|\tilde{\calR}(\tau_r)|}\sum_{j\in\tilde{\calR}(\tau_r)} m_{j}(\tau_r)  
  = \frac{1}{|\tilde{\calR}(\tau_r)|}\sum_{j\in\tilde{\calR}(\tau_r)} r^{j}_{\tau_r}(\theta_1, \theta_2). 
\end{align*}
Let $k_1, \cdots, k_F$ be the indices of the bottom $F$ values that are filtered out by the parameter server. 
Similarly, let $k^{\prime}_1, \cdots, k^{\prime}_F$ be the indices of the top $F$ values that are filtered out by the parameter server. For each $\ell=1, \cdots, F$, there exists $\alpha_{\ell}\in [0, 1]$ such that \footnote{It is worth noting that, in the above equation, for ease of exposition, we drop the time index in the coefficients $\alpha_{\ell}$. }
\[
 \tilde{w}(\tau_r) = \alpha_{\ell}r_{\tau_r}^{k_{\ell}}(\theta_1, \theta_2) + \pth{1-\alpha_{\ell}} r_{\tau_r}^{k^{\prime}_{\ell}}(\theta_1, \theta_2).
\]
Thus, 
\[
 \tilde{w}(\tau_r) = \frac{1}{F} \sum_{\ell=1}^F \pth{\alpha_{\ell}r_{\tau_r}^{k_{\ell}}(\theta_1, \theta_2) + \pth{1-\alpha_{\ell}} r_{\tau_r}^{k^{\prime}_{\ell}}(\theta_1, \theta_2)}.    
\]
We further rewrite $ \tilde{w}(\tau_r)$ as 
\begin{align*}
&r_{t}^{j_{\ell}(t)}(\theta_1, \theta_2) =  \tilde{w}(\tau_r) \\
& = \frac{M-2F}{M} \tilde{w}(\tau_r) + \pth{1- \frac{M-2F}{M}} \tilde{w}(\tau_r) \\
& = \frac{M-2F}{M} \frac{1}{|\tilde{\calR}(\tau_r)|}\sum_{j\in\tilde{\calR}(\tau_r)} r^{j}_{\tau_r}(\theta_1, \theta_2)  \\
&+ \pth{1- \frac{M-2F}{M}} \frac{1}{F} \sum_{\ell=1}^F \pth{\alpha_{\ell}r_{\tau_r}^{k_{\ell}}(\theta_1, \theta_2) + \pth{1-\alpha_{\ell}} r_{\tau_r}^{k^{\prime}_{\ell}}(\theta_1, \theta_2)}\\
& = \frac{1}{M} \sum_{j\in\tilde{\calR}(\tau_r)} r^{j}_{\tau_r}(\theta_1, \theta_2)  \\
&\qquad +  \frac{2}{M} \sum_{\ell=1}^F \pth{\alpha_{\ell}r_{\tau_r}^{k_{\ell}}(\theta_1, \theta_2) + \pth{1-\alpha_{\ell}} r_{\tau_r}^{k^{\prime}_{\ell}}(\theta_1, \theta_2)}. 
\end{align*}
Notably, either $\alpha_{\ell} \ge 1/2$ or $\pth{1 - \alpha_{\ell}} \ge 1/2$. 
Recall that $\abth{\tilde{\calR}(\tau_r)} = M-2F$. 
Hence, we conclude that $\tilde{w}(\tau_r)$ can be written as a convex combination of all the local estimates of the $M$ representatives with at least $M-F$ representatives with weights at least $\frac{1}{M}$. 
By Assumption \ref{ass: sufficiency: Byzantine: non-Bayesian: mixing + information source}, we now that at least one representative from a network in $\calC$ will have corresponding coefficient $\ge \frac{1}{M}$. 
Thus,  when $\theta_1=\theta^*$, by Theorem \ref{thm: convergence: Byzantine: single network}, we have  
\begin{align*}
r_{k_{\tau}}^j(\theta_1, \theta_2) &=  \frac{1}{M} \sum_{j\in\tilde{\calR}(\tau_r)} r^{j}_{\tau_r}(\theta_1, \theta_2)  \\
& \quad +  \frac{2}{M} \sum_{\ell=1}^F \pth{\alpha_{\ell}r_{\tau_r}^{k_{\ell}}(\theta_1, \theta_2) + \pth{1-\alpha_{\ell}} r_{\tau_r}^{k^{\prime}_{\ell}}(\theta_1, \theta_2)}\\
& \ge \frac{1}{M}\beta^{\chi_i(n_i-\phi_i)} D_{KL}^* \pth{k_{\tau}}^2  - \max\{|c_{\min}|, |c_{\max}|\}.  
\end{align*}
Let $\tau\diverge$, we have 
\[
\lim_{\tau\to \infty} r_{k_{\tau}}^j(\theta_1, \theta_2) = +\infty, 
\]
i.e., $\lim\sup_{t\diverge}  r_{k_{\tau}}^j(\theta^*, \theta) = +\infty$. 
For $t\not=\tau_{r}$ for any $r$ and $t\ge t^*$, via the same argument in \cite{vaidya2014iterative}, we are able to write $r_{t}^j(\theta^*, \theta) = \tilde{w}(t)$ as a convex combination of the non-Byzantine representatives of iteration $t$. That is, there exists $\tilde{\alpha}^1_t, \cdots, \tilde{\alpha}^M_t$ such that 
\begin{align*}
 r_{t}^j(\theta^*, \theta) = \tilde{w}(t) = \sum_{i=1}^M \tilde{\alpha}^i_t   r_{t}^{j_i(t)}(\theta^*, \theta). 
\end{align*}
Thus, we have 
\[
r_{t}^j(\theta^*, \theta) \ge - \max\{|c_{\min}|, |c_{\max}|\}. 
\]
Thus, $\lim\inf_{t\diverge} r_{t}^j(\theta^*, \theta) \ge - \max\{|c_{\min}|, |c_{\max}|\}$. Since $\prob{\text{all such }\omega} =1$, we conclude that with probability 1, for all $theta\not=\theta^*$ 
\begin{align*}
&\lim\sup_{t\diverge}  r_{t}^j(\theta^*, \theta) = +\infty, \\
&\lim\inf_{t\diverge} r_{t}^j(\theta^*, \theta) \ge - \max\{|c_{\min}|, |c_{\max}|\}.     
\end{align*}

Similarly, we are able to show that 
with probability 1,  for all $theta\not=\theta^*$ 
\begin{align*}
&\lim\sup_{t\diverge}  r_{t}^j(\theta, \theta^*) \le  \max\{|c_{\min}|, |c_{\max}|\}, \\
&\lim\inf_{t\diverge} r_{t}^j(\theta^*, \theta) =-\infty.    
\end{align*}

It can be easily shown by contradiction (similar to the proof of Lemma \ref{lm: uniqueness}) that if there exists $\tilde{\theta}\in \Theta$ such that for any $\theta\not=\tilde{\theta}$, it holds that 
\[
\lim\sup_{t\diverge}r_t^j(\tilde{\theta}, \theta) = \infty, \qquad \lim\inf_{t\diverge}r_t^j(\tilde{\theta}, \theta) >-\infty, 
\]
and 
\[
\lim\sup_{t\diverge}r_t^j(\theta, \tilde{\theta}) <\infty, \qquad \lim\inf_{t\diverge}r_t^j(\tilde{\theta}, \tilde{\theta}) =-\infty, 
\]
then $\tilde{\theta} = \theta^*$, 
proving Theorem \ref{thm: convergence: Byzantine: general agent}. 
\hfill $\square$

\end{document}